\documentclass[letterpaper]{article} 
\usepackage{aaai24}  
\usepackage{times}  
\usepackage{helvet}  
\usepackage{courier}  
\usepackage[hyphens]{url}  
\usepackage{graphicx} 
\usepackage{natbib}  
\usepackage{caption} 
\usepackage{algorithm}
\usepackage{algorithmic}

\usepackage{newfloat}
\usepackage{listings}
\DeclareCaptionStyle{ruled}{labelfont=normalfont,labelsep=colon,strut=off} 
\floatstyle{ruled}
\newfloat{listing}{tb}{lst}{}
\floatname{listing}{Listing}
\usepackage{subfigure} 
\usepackage{amsmath}
\usepackage{amssymb}
\usepackage{multirow}
\usepackage{makecell}
\usepackage{enumerate}
\usepackage{booktabs}
\newcolumntype{L}[1]{>{\raggedright\let\newline\\\arraybackslash\hspace{0pt}}m{#1}}
\newcolumntype{C}[1]{>{\centering\let\newline\\\arraybackslash\hspace{0pt}}m{#1}}
\newcolumntype{R}[1]{>{\raggedleft\let\newline\\\arraybackslash\hspace{0pt}}m{#1}}

\title{Robust Few-Shot Named Entity Recognition with \\ 
	Boundary Discrimination and Correlation Purification}

\author{
	Xiaojun Xue, Chunxia Zhang\thanks{Corresponding author.}, Tianxiang Xu, Zhendong Niu
}

\affiliations{
	School of Computer Science and Technology, Beijing Institute of Technology, Beijing, China \\
	\{xiaojunx, cxzhang, txxu, zniu\}@bit.edu.cn
}

\usepackage{bibentry}

\begin{document}

\maketitle

\begin{abstract}
	Few-shot named entity recognition (NER) aims to recognize novel named entities in low-resource domains utilizing existing knowledge. 
	However, the present few-shot NER models assume that the labeled data are all clean without noise or outliers, and there are few works focusing on the robustness of the cross-domain transfer learning ability to textual adversarial attacks in Few-shot NER. In this work, we comprehensively explore and assess the robustness of few-shot NER models under textual adversarial attack scenario, and found the vulnerability of existing few-shot NER models. Furthermore, we propose a robust two-stage few-shot NER method with Boundary Discrimination and Correlation Purification (BDCP). Specifically, in the span detection stage, the entity boundary discriminative module is introduced to provide a highly distinguishing boundary representation space to detect entity spans. In the entity typing stage, the correlations between entities and contexts are purified by minimizing the interference information and facilitating correlation generalization to alleviate the perturbations caused by textual adversarial attacks. In addition, we construct adversarial examples for few-shot NER based on public datasets Few-NERD and Cross-Dataset. Comprehensive evaluations on those two groups of few-shot NER datasets containing adversarial examples demonstrate the robustness and superiority of the proposed method.
\end{abstract}

\section{Introduction}
Few-shot named entity recognition (NER) aims to locate and classify new named entities in the target domain where there are only a few labeled examples \cite{LampleBSKD16,KatoAOMSI20,Li00WZTJL22,MaLCZWGZGC23}, which is trained using the present data within the source domain. Recently, few-shot NER has attracted increasing attention, largely due to the fact that it reduces the dependence on labeled data in NER \cite{YangK20,LiCFW22}. Few-shot NER can reflect the generalized learning and cross-domain knowledge transfer abilities of humans \cite{LakeST13,LuJLZ21}, who can infer new knowledge from a few examples based on existing knowledge.

Present few-shot NER methods are usually developed on token-level metric learning \cite{FritzlerLK19,HouCLZLLL20,YangK20} and span level metric learning \cite{YuHZDPL21,WangXLZCCS22}. In the former family, novel entities in target domain are recognized by measuring the distance between each query token and the prototype of each entity category or each token of support examples \cite{SnellSZ17}. In contrast, the latter bypass the token-wise label dependency issue by measuring the distance between spans \cite{decomposed}. However, existing few-shot NER methods typically assume that the examples are all clean without noise or outliers \cite{LuJLZ21}, which is overly idealistic in the real world. Thus, it is essential to maintain robust cross-domain transfer learning ability when handling adversarial examples with interferences for few-shot NER. In addition, at present, some works have investigated the robustness of NER task \cite{LinG0M021,WangDXZZG0CH22}, emphasizing the context-based reasoning. Different from NER task, few-shot NER focuses on learning the correlation transfer between entity and contexts from source domain to target domain. Our deep investigation on literature reveals that this particular topic has not gained enough attention.

\begin{figure}[tpb]
	\center{\includegraphics[width=7.5cm]  {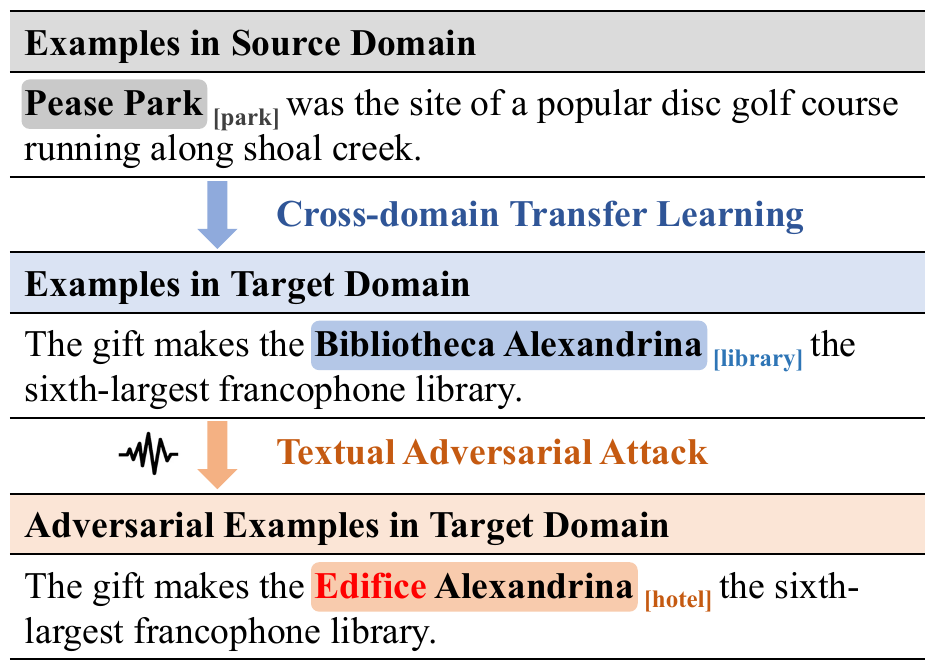}} 
	\caption{\label{fig_example} Vulnerabilities exhibited in few-shot NER under textual adversarial attack (i.e. synonym substitution) scenario. The subscripts indicate entity types.} 
\end{figure}

Therefore, in this paper, we explore and assess the adversarial robustness of the cross-domain transfer learning ability in few-shot NER. For textual adversarial attack, synonym substitution is a widely used method, where the words in original texts are replaced by their synonyms \cite{LiMGXQ20,LiXZLZZCH21,ZengXZH23}. The substituted words are imperceptible to humans, while they can “fool” the neural networks to make wrong predictions. In this work, synonym substitution attack is performed on clean samples in few-shot NER. We have observed that synonym substitution attack misleads the existing few-shot NER models, as shown in Figure \ref{fig_example}, and resulting in a significant drop in recognition performance for unseen entity types (detailed in Experiments section). Typically, the cross-domain transfer learning ability in few-shot NER is vulnerable when handling adversarial examples.

Motivated by the above observations, we proposed a robust two-stage few-shot NER method with Boundary Discrimination and Correlation Purification (BDCP). Overall, the span detection stage is to detect the entity spans in the input text, then the entity typing stage is to classify the detected entity spans to the corresponding unseen entity types. First, in the span detection stage, the entity boundary discriminative module is introduced to provide a highly distinguishing boundary representation space to detect entity spans, which contains multiple components assigned to all boundary classes. The span detection is regarded as a boundary classification task (e.g. BIOES \cite{decomposed}), and the token representations are diversely assigned to the corresponding closest components in the entity boundary discriminative module. The backbone model (i.e. encoder layer) and the boundary discriminative module are simultaneously learned by utilizing two mutually complementary losses, which can improve the adversarial robustness of span detection. Second, in the entity typing stage, the correlations between entities and contexts are purified to alleviate the perturbations caused by adversarial attacks. The correlations are purified by minimizing the interference information in correlations and facilitating correlation generalization from an information theoretic perspective, which can alleviate the perturbations caused by textual adversarial attacks.

The contributions of this paper are summarized as follows:
\begin{itemize}
	\item We explore and assess the adversarial robustness of the cross-domain transfer learning ability in few-shot NER for the first time. A robust two-stage few-shot NER method with Boundary Discrimination and Correlation Purification (BDCP) is proposed to defend against the textual adversarial attacks. The codes are publicly available\footnote{https://github.com/ckgconstruction/bdcp}.
	
	\item Entity boundary discriminative module is introduced to provide a highly distinguishing boundary representation space, thereby improving the adversarial robustness of entity span detection. Two mutually complementary losses are utilized to diversely assign each token representation to corresponding component.
	
	\item Aiming at improving the adversarial robustness of entity typing, we implement correlation purification between entities and contexts by minimizing the interference information in correlations and facilitating correlation generalization. Correlation purification can alleviate the perturbations caused by textual adversarial attacks.
	
\end{itemize}

\section{Related Work}
\subsection{Few-Shot Named Entity Recognition}
Current works about few-shot named entity recognition (NER) mainly focus on metric learning methods at token-level \cite{FritzlerLK19,YangK20} and span-level \cite{decomposed,WangXLZCCS22}. The token-level approaches recognize novel entities in query samples by measuring the distance between each query token and the prototype of each entity category or each token of support examples \cite{SnellSZ17}, while the span-level methods bypass the token-wise label dependency issue by measuring the distance between spans.

In the first family, \citeauthor{FangWMXHJ23} \shortcite{FangWMXHJ23} designed an additional memory module that stored token representations of entity types to adaptively learn cross-domain entities.
In the second family, \citeauthor{decomposed} \shortcite{decomposed} introduced a decomposed meta-learning approach to decompose the few-shot NER task into span detection and entity typing, which enabled the few-shot NER model to learn suitable initial parameters and embedding space. 

However, the existing few-shot NER methods suffer a significant drop in recognition performance for unseen entity types when handling adversarial examples. It demonstrates the vulnerability of the cross-domain transfer learning abilities of present methods under textual adversarial attack scenarios.

\subsection{Adversarial Robustness on Texts}

Recently, a magnitude of adversarial attacks have been introduced for texts \cite{ZhouJCW19,LiuZRLLCQG0H22,ZengXZH23}, such as synonym substitution \cite{LinG0M021}, adversarial perturbation generating \cite{Wang0D021,ZhuCGSGL20}, which can maintain human understanding of sentences but “fool” the deep neural networks. 
Textual adversarial attacks make deep neural networks output incorrect predictions and point out the vulnerability of current models regarding textual adversarial attacks. 
Aiming at verifying the robustness of the NER models, \citeauthor{LinG0M021} \shortcite{LinG0M021} generated entities substitutions using the other entities of the same semantics in Wikidata, and utilized pre-trained language models \cite{bert} to replace the words in context. 
\citeauthor{ZengXZH23} \shortcite{ZengXZH23} generated a set of copies for input texts by randomly masking words to improve robustness.

Current adversarial robustness tasks on text mainly focus on learning ability in the same domain, such as text classification \cite{LiMGXQ20,0008X0XZMC0Z23}, sentiment analysis \cite{LiXZLZZCH21} and named entity recognition \cite{LinG0M021}. However, there are few works about  the robustness of cross-domain transfer learning ability for texts. 
In this work, we focus on the effect of textual adversarial attacks on the cross-domain transfer learning ability in few-shot NER task.

\begin{figure*}[!htb] 
	\centering
	\includegraphics[width=15.6cm]  {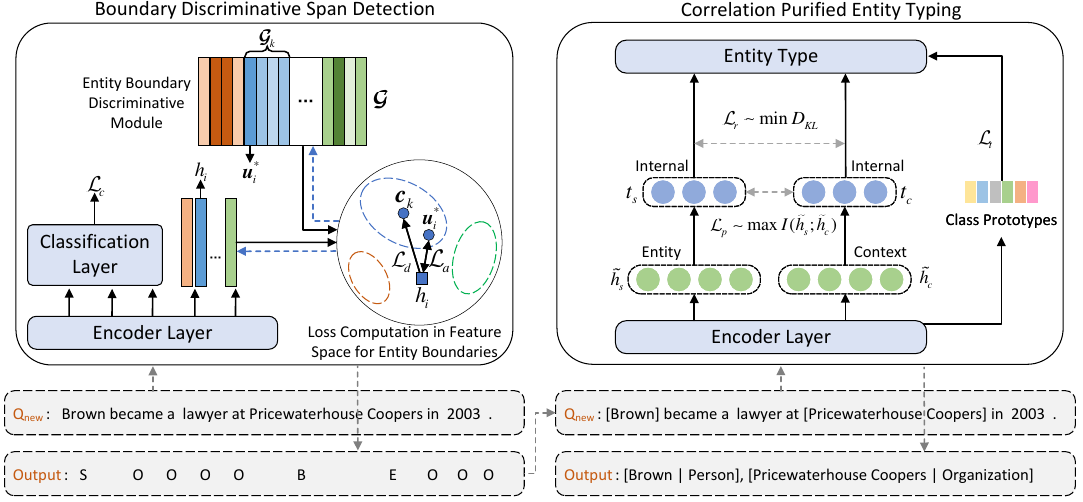}
	\caption{Overview of the robust two-stage few-shot NER method with Boundary Discrimination and Correlation Purification (BDCP). In the span detection stage, we introduce entity boundary discriminative module $\boldsymbol{\mathcal{G}}$ to provide a highly distinguishing boundary representation space. Each token representation $h_i$ is diversely assigned to the closest component $\boldsymbol{u}_{i}^{*}$ through two mutually complementary assignment loss $\mathcal{L}_a$ and diversity loss $\mathcal{L}_d$. The blue dashed lines represent backpropagation. In the entity typing stage, the correlations between entities and contexts are purified to alleviate the perturbations caused by textual adversarial attacks. Correlation purification is implemented by minimizing interference information in correlations (i.e. $\mathcal{L}_r$) and facilitating correlation generalization (i.e. $\mathcal{L}_p$).} 
	\label{fig_3_1}
\end{figure*}

\section{Preliminaries}
\noindent \textbf{Few-Shot Named Entity Recognition.}
Given a sequence $\boldsymbol{x}=\{{{x}_{i}}\}_{i=1}^{n}$ containing $n$ tokens, named entity recognition (NER) task aims to output the entity sequences $\boldsymbol{e}={{\{{{e}_{j}}\}}_{j\ge 0}}$ and assign the corresponding entity type ${y}_{j}$ to each entity sequence ${e}_{j}$. 
The few-shot NER model is trained in a source domain ${{\boldsymbol{\varepsilon}}_{source}}=\{({\mathcal{S}_{s}},{\mathcal{Q}_{s}},{\mathcal{Y}_{s}})\}$, where ${\mathcal{S}_{s}},{\mathcal{Q}_{s}},{\mathcal{Y}_{s}}$ represent the support set, query set, and the entity types in the training data, respectively. 
Then the few-shot NER model is transferred to a data-scarce target domain ${{\boldsymbol{\varepsilon}}_{target}}=\{({\mathcal{S}_{t}},{\mathcal{Q}_{t}},{\mathcal{Y}_{t}})\}$ for testing with the similar data construction. 
A few-shot NER model learned in the source domain ${\boldsymbol{\varepsilon}}_{source}$ is expected to leverage the support set $\mathcal{S}_{t}$ of the target domain to predict novel entities in the query set $\mathcal{Q}_{t}$. Since ${\mathcal{Y}_{s}}\cap{\mathcal{Y}_{t}}=\emptyset$, the few-shot NER model needs to learn cross-domain transfer knowledge to be generalized to unseen entity types in few-shot NER. 
In the $N$-way $K$-shot setting, there are $N$ entity types in the target domain (i.e. $|{\mathcal{Y}_{t}}|=N$), and each entity type is associated with $K$ examples in the support set ${\mathcal{S}_{t}}$.

\noindent \textbf{Adversarial Attack on Few-Shot NER.} 
The entities and corresponding entity types recognized by the few-shot NER model in the target domain are denoted as the set $\{{{C}_{i}}\}$, where ${{C}_{i}}=({{e}_{p}},{{y}_{p}})$, ${{e}_{p}}$ and ${y}_{p}$ are the entity and corresponding entity type. 
The adversarial attack on few-shot NER aims to construct adversarial examples to ``fool" the neural network based few-shot NER models. 
An adversarial example $\boldsymbol{{x}'}$ makes the few-shot NER model that correctly recognizes entities and corresponding entity types to predict incorrect entity-type pairs $\{{{C}'_{i}}\}$, i.e.
\begin{equation}
	\{{{C}_{i}}\}\neq\{{{C}'_{i}}\}.
	\label{equ_3_1}
\end{equation}
The adversarial example $\boldsymbol{{x}'}$ is constructed by replacing the original tokens with synonyms ${w}'_i$ in the synonym set.

\noindent \textbf{Information Bottleneck (IB)} principle utilizes the idea of mutual information to analyze the training and inference of deep neural networks \cite{Shwartz-ZivT17,TishbyZ15}. Given the input data $X$ and the label $Y$, it attempts to learn an internal representation $T$ that makes an information trade-off between predictive accuracy and representation compression:
\begin{equation}
	{\mathcal{L}_{IB}}=-I(T;Y)+\beta *I(T;X),
	\label{equ_3_1_2}
\end{equation}
\noindent where $I$ stands for mutual information (MI), aiming to measure the interdependence between two variables. $\beta$ is the Lagrange multiplier that controls the trade-off between two MI terms. By optimizing the loss $\mathcal{L}_{IB}$, the IB principle compresses the noise data in $X$ while retaining enough features in $X$ to predict $Y$.

\section{Methodology}

\subsection{Overview of the Proposed Method}
Technically, we found that the decline in cross-domain transfer learning ability for few-shot NER is reflected in: 1) entity span detection errors; 2) inaccurate unseen entity type classification. To solve the above problems, we propose a robust two-stage few-shot NER method with Boundary Discrimination and Correlation Purification (BDCP). Figure \ref{fig_3_1} illustrates the architecture of the proposed BDCP method. Following \citeauthor{decomposed} \shortcite{decomposed}, the few-shot NER task is decomposed into span detection and entity typing. Overall, the span detection stage is designed to detect the entity spans in the input text, then those entity spans are classified into corresponding unseen entity types in the entity typing stage.

Different from \citeauthor{decomposed} \shortcite{decomposed}, in the span detection stage, for the purpose of improving the adversarial robustness of entity span detection, entity boundary discriminative module is introduced to provide a highly discriminative boundary representation space. In the entity typing stage, aiming at improving the adversarial robustness of entity typing, the correlations between entities and contexts are purified by minimizing the interference information and facilitating correlation generalization. Note that, the span detection stage and the entity typing stage are learned serially. The proposed BDCP method does not introduce additional computational cost like adversarial training and data augmentation.

\subsection{Adversarial Examples Generating}
Textual adversarial attack is conducted on the examples in the support set ${\mathcal{S}_{t}}$ and the query set ${\mathcal{Q}_{t}}$ of the target domain. The textual adversarial attack algorithm BERT-Attack \cite{LiMGXQ20} is used to perform synonym substitution and generate adversarial examples. \footnote{More details about the attack algorithm and generation implementation are listed in Appendix.} 
The cross-domain transfer learning ability is reflected through the performance of the few-shot NER model in the query set ${\mathcal{Q}_{t}}$ of the target domain. 
Textual adversarial attacks in the real world usually exist randomly in textual data, hence we generate adversarial examples against the entire original examples without distinguishing entities and contexts \cite{LinG0M021}.

\subsection{Boundary Discriminative Span Detection}
Span detection is regarded as a token-level label classification process. The entity boundary discriminative module $\boldsymbol{\mathcal{G}}$ is introduced to alleviate the problem of entity boundary detection errors, which provides an additional robust representation space for entity span boundaries. 
Entity boundary discriminative module $\boldsymbol{\mathcal{G}}$ contains $N_b$ components blocks $\boldsymbol{\mathcal{G}}_k$, where $N_b$ denotes the number of entity boundary classes. ${{G}_{k}}=\left\{ {{u}_{i}} \right\}_{i=1}^{{{N}_{c}}}$ is the set of components ${{u}_{i}}\in {\mathbb{R}^{C}}$ assigned to boundary class $k$, where $N_c$ represents the number of components assigned to each boundary class. 

\noindent \textbf{Boundary Assignment.}
First, the adversarial example $\boldsymbol{{x}'}=\{{{w}_{i}}\}_{i=1}^{n}$ is input into an encoder $f_\theta$ to generate token representations $\boldsymbol{h}=\{{{h}_{i}}\}_{i=1}^{n}$:
\begin{equation}
	\boldsymbol{h}=f_\theta(\boldsymbol{{x}'}).
	\label{equ_3_2}
\end{equation}

Each token representation $h_i$ is matched amongst the elements in each components block $\boldsymbol{\mathcal{G}}_k$ according to the cosine similarity, thus obtaining the closest component $\boldsymbol{u}_{i}^{*}$ corresponding to $h_i$:
\begin{equation}
	\boldsymbol{u}_{i}^{*}=\underset{{{u}_{j}}\in {\boldsymbol{\mathcal{G}}_k}}{\mathop{\arg \max }}\,\frac{{{h}_{i}}\centerdot {{u}_{j}}}{||{{h}_{i}}||\text{ }||{{u}_{j}}||},
	\label{equ_3_4}
\end{equation}
\noindent where $\centerdot$ represents dot product. 
Inspired by the mixture based feature space \cite{AfrasiyabiLG21}, we improve the angular margin-based softmax function \cite{DengGXZ19} with a temperature variable $\tau$:
\begin{equation}
	\begin{aligned}
		&{{p}_{\theta }}({{v}_{j}}|{{h}_{i}},\boldsymbol{\mathcal{G}})= \\
		&\frac{{{e}^{\cos (\angle ({{h}_{i}},{{u}_{j}})+m)/\tau }}}{{{e}^{\cos (\angle ({{h}_{i}},{{u}_{j}})+m)/\tau }}+\sum\limits_{{{u}_{l}}\in \{\boldsymbol{\mathcal{G}}\backslash {{u}_{j}}\}}{{{e}^{\cos (\angle ({{h}_{i}},{{u}_{l}}))/\tau }}}},
		\label{equ_3_3}
	\end{aligned}
\end{equation}
\noindent where $\angle ({{h}_{i}},{{u}_{j}})=arccos({h}_{i}^\top{{u}_{i}}/(||{h}_{i}||||{{u}_{j}}||))$, $v_j$ denotes the pseudo-label associated to $u_j$, and $m$ represents a margin.

Next, every token representation $h_i$ is assigned to the closest component $\boldsymbol{u}_{i}^{*}$ in $\boldsymbol{\mathcal{G}}$ utilizing the assignment loss ${\mathcal{L}_{a}}$:
\begin{equation}
	\mathcal{L}_a=-\frac{1}{L}\sum\limits_{i=1}^{L}{\log {{p}_{\theta }}({{v}_{i}^*}|{{h}_{i}},\boldsymbol{\mathcal{G}})},
	\label{equ_3_5}
\end{equation}
\noindent where $L$ represents the number of all tokens in a batch, and ${v}_{i}^*$ denotes the one-hot pseudo-label corresponding to ${u}_{i}^*$.

\noindent \textbf{Diverse Assignment.}
Each entity boundary class is supposed to be mapped to multiple components rather than a single one in $\boldsymbol{\mathcal{G}}_k$, which can improve the adversarial robustness. Because in this way, more generalized boundary representation space can be leveraged to detect the entity boundaries. 
Training the backbone model and the entity boundary augmentation matrix only on the assignment loss $\mathcal{L}_a$ usually results in that the tokens of the entity boundary class $k$ are  assigned to a single component ${u_i}\in{\mathcal{G}_k}$. To avoid this, the diversity loss $\mathcal{L}_d$ is designed to facilitate the diversity of component selection for every entity boundary class.

The diversity loss $\mathcal{L}_d$ pushes the token representation $h_i$ towards the centroid of components associated with its entity boundary class. The centroid ${c}_{k}$ for entity boundary class $k$ is defined as:
\begin{equation}
	{\boldsymbol{c}_{k}}=(1/|{\mathcal{G}_{k}}|)\sum\nolimits_{{{u}_{j}}\in {\mathcal{G}_{k}}}{{{u}_{j}}},
	\label{equ_3_6}
\end{equation}
\noindent where $|\mathcal{G}_{k}|$ denotes the number of components for each entity boundary labels $y_i$. 
For the centroids set $\boldsymbol{C}={\{c_k\}}_{k=1}^{L_e}$, the diversity loss $\mathcal{L}_d$ is calculated as:
\begin{equation}
	{\mathcal{L}_{d}}=-\frac{1}{L}\sum\limits_{i=1}^{L}{\log {{p}_{\theta }}({{y}_{i}}|{{h}_{i}},sg[\boldsymbol{C}])},
	\label{equ_3_7}
\end{equation}
\noindent where $sg$ denotes stopgradient that protects specific variables from backpropagation. It prevents the components of each boundary class from collapsing into a single point. 

Following \citeauthor{decomposed} \shortcite{decomposed}, we also use averaged cross-entropy loss with a maximum term as training loss:
\begin{equation}
	\begin{aligned}
		{\mathcal{L}_{c}}=& \frac{1}{L}\sum\limits_{i=1}^{L}CrossEntropy({{y}_{i}},p({{w}_{i}})) \\
		& +\alpha \underset{i\in \{1,2,...,L\}}{\mathop{\max }}\,CrossEntropy({{y}_{i}},p({{w}_{i}})),
		\label{equ_3_9}
	\end{aligned}
\end{equation}

\noindent where the maximum term is leveraged to alleviate insufficient learning for tokens with higher loss. $p$ represents the fully connected layer with $softmax$ activation function. $\alpha$ is the weighting factor.

\noindent \textbf{Final Objective.} 
Overall, the training loss in boundary discriminative span detection stage is the combination of Eq. (\ref{equ_3_5}), Eq. (\ref{equ_3_7}) and Eq. (\ref{equ_3_9}):
\begin{equation}
	{\mathcal{L}_{sp}}={\mathcal{L}_{c}}+\gamma_1 {\mathcal{L}_{a}}+\gamma_2 {\mathcal{L}_{d}},
	\label{equ_3_10}
\end{equation}
\noindent where $\gamma_1$ and $\gamma_2$ are weighting factors for assignment loss and diversity loss, respectively.

\subsection{Correlation Purified Entity Typing}
In the entity typing stage, the entity spans obtained in the span detection stage are classified into corresponding entity types. 
In this paper, the correlations refer to the correlations between entities and contexts for simplicity.

Textual Adversarial attacks bring interferences to the correlations, which has an adverse effect on cross-domain transfer learning. Therefore, we design correlation purified entity typing to filter the interference information existing in the correlations and facilitate the correlation generalization, which can alleviate the perturbations caused by textual adversarial attacks. 
Inspired by \citeauthor{TishbyZ15} \shortcite{TishbyZ15} and \citeauthor{WangDXZZG0CH22} \shortcite{WangDXZZG0CH22}, we explicitly minimize the interference information and maximize the interactive information between entities and contexts. 

For entity typing, the adversarial example ${x}’$ is input into another encoder ${g}_{\theta }$ to generate token representations $\boldsymbol{\tilde{h}}=\{{{\tilde{h}}_{i}}\}_{i=1}^{n}$:
\begin{equation}
	\boldsymbol{\tilde{h}}={{g}_{\theta }}({x}').
	\label{equ_3_g}
\end{equation}

The positions of the entity spans can be obtained from the span detection stage. Subsequently, the entity span representation ${\tilde{h}}_{s}$ is computed by averaging all token representations inside the entity span. 
The context representation ${\tilde{h}}_{c}$ is computed by averaging the rest of the token representations in ${x}'$ other than the entity spans.

Intuitively, for few-shot NER in the textual adversarial scenarios, the correlations between entities and contexts are essential to predict unseen entities, and the interferences in adversarial examples are interfering features. Motivated by the chain rule of mutual information and the contrastive strategy \cite{Federici0FKA20,WangDXZZG0CH22}, the mutual information $I(x;{{\tilde{h}}_{s}})$ between entity span $\tilde{h}_s$ and original example $x$ can be decomposed into two parts:
\begin{equation}
	I(x;{{\tilde{h}}_{s}})=\underbrace{I({{{\tilde{h}}}_{s}};{{x}'})}_{predictable}+\underbrace{I(x;{{{\tilde{h}}}_{s}}|{{x}'})}_{specific},
	\label{equ_3_12}
\end{equation}

\noindent where $x'$ denotes to the adversarial example. 
$I({{\tilde{h}}_{s}};{{x}'})$ refers to the information in $\tilde{h}_s$ that is predictable for ${x}'$, i.e. non-entity-specific information. 
$I(x;{{\tilde{h}}_{s}}|{{x}'})$ indicates the information in $\tilde{h}_s$ that is unique to $x$ but is unpredictable for ${x}'$, i.e. entity-specific information.

Consequently, the entity-specific information is interfering, and any token representation $h$ that encompasses the information jointly shared by $x$ and ${x}'$ would also contain the requisite label information. Then Eq. (\ref{equ_3_12}) can be approximated as follows:
\begin{equation}
	maximize\ I({\tilde{h}_s};y)\sim I({\tilde{h}_s};{x}'),
	\label{equ_3_13}
\end{equation}
\begin{equation}
	minimize\ I({x};{\tilde{h}_s}|y)\sim I({x};{\tilde{h}_s}|{x}'),
	\label{equ_3_14}
\end{equation}

\noindent \textbf{Correlation Facilitating.} 
To facilitate correlation generalization between entities and contexts, Eq. (\ref{equ_3_13}) is utilized to maximize the generalization information of textual representations. Similar to \citeauthor{WangDXZZG0CH22} \shortcite{WangDXZZG0CH22}, it has been proved that $I(\tilde{h}_s,\tilde{h}_c)$ is a lower bound of $I(\tilde{h}_s,{x}')$ (proof is detailed in Appendix). InfoNCE \cite{infonce} can be leveraged to approximate $I(\tilde{h}_s,\tilde{h}_c)$, which is a lower bound of mutual information. 
It can be known from previous work \cite{infonce} that when the number of tokens is lager, the lower bound is closer to InfoNCE. During training, minimizing the InfoNCE loss can maximize the lower bound of mutual information. Thus, the goal of facilitating correlation generalization is optimized by:
\begin{equation}
	{\mathcal{L}_{p}}=-{\mathbb{E}_{1}}\left[ {{g}_{p}}({{{\tilde{h}}}_{s}},{{{\tilde{h}}}_{c}})-{\mathbb{E}_{2}}(\log \sum\limits_{{{{\tilde{h}}}_{i}}\in {{{\tilde{h}}}_{c}}}{\exp ({{g}_{p}}({{{\tilde{h}}}_{s}},{{{\tilde{h}}}_{c}}))}) \right],
	\label{equ_3_lp}
\end{equation}
\noindent where $\mathbb{E}_{1}$ and $\mathbb{E}_{2}$ are two different activation functions. 
${g}_{p}$ is the compatibility scoring function implemented by a linear neural network.

\noindent \textbf{Correlation Purification.} 
Aiming at alleviating the adverse effects of interferences in adversarial examples, Eq. (\ref{equ_3_14}) is exploited to minimize the interference information. To this end, we minimize an upper bound of $I({x};{\tilde{h}_s}|{{x}'})$ (proof is detailed in Appendix), which is formulated as:
\begin{equation}
	{{p}_{ib}}(t|h)=\mathcal{N}(t|f_{ib}^{\mu }(h),f_{ib}^{M}(h))
	\label{equ_3_pib}
\end{equation}
\begin{equation}
	{{\mathcal{L}}_{r}}={{D}_{KL}}[{{p}_{ib}}({{t}_{s}}|{{\tilde{h}}_{s}})||{{p}_{ib}}({{t}_{c}}|{{\tilde{h}}_{c}}))]
	\label{equ_3_lr}
\end{equation}
\noindent where $p_{ib}$ denotes the information bottleneck layer, 
$t$ refers to internal representation, 
$f_{ib}^{\mu }$ and $f_{ib}^{v}$ are multilayer perceptrons (MLP) that compute the mean ${\mu}$ and the covariance matrix $M$ of $t$, respectively, 
and $\mathcal{N}$ stands for the reparameterization trick \cite{KingmaW13}. 
${D}_{KL}$ refers to Kullback-Leibler divergence.

\noindent \textbf{Final Objective.} 
We utilize ProtoNet \cite{SnellSZ17} to calculate class prototypes and use cross-entropy loss to compute prototype loss $\mathcal{L}_{t}$, same as \citeauthor{decomposed} \shortcite{decomposed}. 
Due to space limitation, the calculation processes are listed in Appendix. 
Finally, the training loss in correlation purified entity typing stage is the combination of $\mathcal{L}_{t}$, $\mathcal{L}_{p}$ and $\mathcal{L}_{r}$:
\begin{equation}
	{\mathcal{L}_{sp}}={\mathcal{L}_{t}}+\gamma_3 {\mathcal{L}_{p}}+\gamma_4 {\mathcal{L}_{r}},
	\label{equ_3_l2}
\end{equation}

\noindent where $\gamma_3$ and $\gamma_4$ are weighting factors for the correlation facilitation loss and interference information filtering loss, respectively. In addition, we adopt the same meta-learning strategy as \citeauthor{decomposed} \shortcite{decomposed} to train and evaluate the model in two stages.

\begin{table*}[!h]
	\newcommand{\tabincell}[2]{\begin{tabular}{@{}#1@{}}#2\end{tabular}}
	\centering
	\caption{Performance on Few-NERD.}
	\label{table_4_1}
	\begin{tabular}{lc|c|c|c|c|c|c|c}
		\toprule 
		\multirow{3}{*}{Models } & \multicolumn{4}{c}{INTRA} & \multicolumn{4}{c}{INTER}\\
		\cline{2-9}
		& \multicolumn{2}{c|}{5-1} & \multicolumn{2}{c|}{10-5} & \multicolumn{2}{c|}{5-1} & \multicolumn{2}{c}{10-5} \\
		\cline{2-9}
		& Clean & Attack & Clean  & Attack & Clean & Attack & Clean  & Attack  \\ 
		\midrule 
		ProtoBERT (Fritzler et al. 2019)  & 23.45 & 12.45 & 34.61 & 29.08 & 44.44 & 38.30 & 53.97 & 45.23  \\ 
		StructShot (Yang et al. 2020) & 35.92 & 21.62 & 26.39 & 19.71 & 57.33 & 42.78 & 49.39 & 42.75 \\
		CONTAINER \cite{DasKPZ22} & 40.43 & 24.58 & 47.49 & 40.15 & 55.95 & 41.45 & 57.12 & 48.19 \\
		Decomposed \cite{decomposed} & 52.04 & 33.73 & 56.84 & 45.16 & 68.77 & 49.36 & 68.32 & 54.68  \\ 
		\midrule
		RockNER \cite{LinG0M021} & 52.17 & 35.29 & 57.25 & 46.50 & 68.92 & 51.54 & 68.43 & 57.64 \\ 
		RanMASK \cite{ZengXZH23} & 52.33 & 36.16 & 56.79 & 44.18 & 67.58 & 52.37 & 68.61 & 58.86 \\
		\midrule
		Ours & \textbf{52.63} & \textbf{40.76} & \textbf{57.46} & \textbf{50.71} & \textbf{69.59} & \textbf{56.84} & \textbf{68.87} & \textbf{64.97}  \\ 
		\bottomrule 
	\end{tabular}
\end{table*}

\begin{table*}[!h]
	\newcommand{\tabincell}[2]{\begin{tabular}{@{}#1@{}}#2\end{tabular}}
	\centering
	\caption{Performance on Cross-Dataset.}
	\label{table_4_2}
	\begin{tabular}{lc|c|c|c|c|c|c|c}
		\toprule 
		\multirow{3}{*}{Models } & \multicolumn{8}{c}{5-shot} \\
		\cline{2-9}
		& \multicolumn{2}{c|}{News} & \multicolumn{2}{c|}{Wiki} & \multicolumn{2}{c|}{Social} & \multicolumn{2}{c}{Mixed} \\
		\cline{2-9}
		& Clean & Attack & Clean  & Attack & Clean & Attack & Clean  & Attack  \\ 
		\midrule 
		MatchingNet \cite{VinyalsBLKW16} & 19.85 & 13.95 & 5.58 & 4.52 & 6.61 & 5.13 & 8.08 & 6.58 \\
		SimBERT \cite{HouCLZLLL20} & 32.01 & 25.56 & 10.63 & 8.27 & 8.20 & 6.74 & 21.14 & 14.62  \\ 
		L-TapNet+CDT \cite{HouCLZLLL20} & 45.35 & 31.23 & 11.65 & 9.69 & 23.30 & 15.38 & 20.95 & 13.44 \\
		Decomposed \cite{decomposed} & 58.18 & 42.87 & 31.36 & 23.39 & 31.02 & 22.46 & 45.55 & 34.19  \\ 
		\midrule
		RockNER \cite{LinG0M021} & 58.32 & 44.51 & 31.63 & 24.45 & 31.52 & 24.18 & 45.78 & 36.48 \\ 
		RanMASK \cite{ZengXZH23} & 58.54 & 45.19 & 32.06 & 22.64 & 31.67 & 24.69 & 44.35 & 37.51 \\
		\midrule
		Ours & \textbf{58.76} & \textbf{51.36} & \textbf{32.17} & \textbf{27.91} & \textbf{31.84} & \textbf{27.29} & \textbf{46.29} & \textbf{40.73}  \\ 
		\bottomrule
	\end{tabular}
\end{table*}

\section{Experiments}

\subsection{Experimental Settings}

\noindent \textbf{Datasets.} 
Comprehensive experiments are conducted on two groups of datasets to evaluate the adversarial robustness of the proposed method: (1) \textbf{Few-NERD} \cite{DingXCWHXZL20} contains 8 coarse-grained and 66 fine-grained entity types with hierarchical structure for few-shot NER\footnote{https://github.com/thunlp/Few-NERD}, addressing two tasks named INTRA and INTER. Specifically, entities in the train/dev/test splits belong to different coarse-grained entity types in INTRA, while fine-grained entity types are mutually disjoint and the coarse-grained entity types are shared in INTER. (2) four sub-datasets from four domains are used in \textbf{Cross-Dataset} \cite{HouCLZLLL20}: News (CoNLL-2003) \cite{Sang02}, Wiki (GUM) \cite{Zeldes17}, Social (WNUT-2017) \cite{DerczynskiNEL17}, Mixed (Ontonotes) \cite{PradhanMXNBUZZ13}\footnote{https://github.com/AtmaHou/FewShotTagging}. 

For the Few-NERD dataset, the episodes in \citeauthor{decomposed} \shortcite{decomposed} are used, where each episode contains one N-way K-shot few-shot NER task. Totally, 20,000/1,000/5,000 episodes are employed for training/validation/testing. Accordingly, 5-way 1-shot and 10-way 5-shot setups are utilized for INTRA and INTER. 
For the Cross-Dataset, two sub-datasets are used to construct training episodes. One sub-dataset and one of the remaining sub-datasets are utilized to construct validation episodes and test episodes, respectively. In the 5-shot setup on Cross-Dataset, 200/100/100 episodes are used for training/validation/testing, as in \citeauthor{HouCLZLLL20} \shortcite{HouCLZLLL20}. The sub-datasets for training, validation and test are mutually disjoint from each other, so that all the models can be tested in cross-domain scenarios.

\noindent \textbf{Baselines.} 
The typical metric-learning based baselines are considered first.
For Few-NERD, the proposed method is compared with typical Few-shot NER methods ProtoBERT \cite{FritzlerLK19}, StructShot \cite{YangK20}, CONTAINER \cite{DasKPZ22} and Decomposed \cite{decomposed}. 
For Cross-Dataset, the proposed method is compared with SimBERT \cite{HouCLZLLL20}, L-TapNet+CDT \cite{HouCLZLLL20} and Decomposed \cite{decomposed}. 
Second, since there are currently few robust few-shot NER methods, the direct adaptations of the following textual defense methods are also compared with our BDCP model: RockNER \cite{LinG0M021}, RanMASK \cite{ZengXZH23} for both Few-NERD and Cross-Dataset. More details about the baselines are listed in Appendix. We report the F1 scores of the models handling clean and adversarial examples.

\noindent \textbf{Implementation Details.} 
Two separate BERT-base-uncased models \cite{bert} are used to implement encoders $f_\theta$ and $g_\theta$. 
The batch size, dropout probability and maximum sequence length are set to 64, 0.2, 128, respectively. 
The model is trained using the AdamW optimizer \cite{LoshchilovH19}, and the initial learning rates for two stages are set to 3e-5 and 1e-4, respectively. 
The temperature variable $\tau$ and the margin $m$ are 0.025 and 0.01. The number of components for each entity boundary class in $\boldsymbol{\mathcal{G}}$ is 15. The weighting factors $\gamma_1$, $\gamma_2$, $\gamma_3$ and $\gamma_4$ are set to 0.1, 0.1, 1e-3,1e-5, respectively. 
For clean examples, the results of typical few-shot NER models are reported from \citeauthor{decomposed} \shortcite{decomposed}, and we reproduce the rest of the models to report results. For adversarial examples, we reproduce all the baseline models to obtain results.

\subsection{Experimental Results}
\noindent \textbf{Performance Comparison.} 
Table \ref{table_4_1} and Table \ref{table_4_2} report the overall comparison results of different models on Few-NERD and Cross-Dataset datasets, under both Clean and Attack scenarios. Here we can observe that:

\begin{enumerate}[1)]
	
	\item All baseline models suffer from a significant drop in performance under textual adversarial attack, indicating the vulnerability of the cross-domain transfer learning ability for existing few-shot NER methods. For example, the advanced Decomposed model dramatically drops an average of 15.76 and 10.80 points on the Few-NERD and Cross-Dataset datasets, respectively. The reason lies in that those models learn the preferences and usage patterns for specific words during training, and are fooled by the interferences brought by textual adversarial attacks.
	
	\item RockNER and RanMASK improve robustness through data augmentation and random masking respectively. 
	Compared with the typical Decomposed model, the performance of RockNER is slightly improved, which shows that increasing data scale is conducive to improving the robustness of few-shot NER. However, random masking may reduce the useful information where there is a lack of labeled samples. Hence RanMASK cannot improve the cross-domain transfer learning ability in that case (e.g. $10$-way $5$-shot setting in INTRA).
	
	\item By utilizing entity boundary discrimination and correlation purification, the proposed BDCP model outperforms all the baseline methods in both Clean and Attack scenarios. Under textual adversarial attack scenario, compared with the typical Decomposed model, the performance of our BDCP model is improved by 7.59\% and 6.10\% in terms of F1 on the Few-NERD and Cross-Dataset datasets, respectively. The advantages of the BDCP model in adversarial robustness lie in that: (a) the boundary discrimination module improves the robustness of span detection by providing highly distinguishing and diverse entity boundary representation space; (b) the correlation purification module mitigates the interference of attacks by restraining unpredictable information and the KL divergence of the internal representations, entities and contexts.
	
\end{enumerate}

\noindent \textbf{Ablation Study}.
We design ablation studies to analyze the effects of different modules on adversarial robustness in the BDCP method. 
Table \ref{table_4_3} shows the results in different settings, taking the Few-NERD dataset containing adversarial examples as an example. 
``Base" model refers to Decomposed model \cite{decomposed}.
``+components" indicates that both assignment loss and diversity loss are added. 
``+purify" means simultaneously using correlation facilitation loss and interference information filtering loss. 
It can be observed that the ``+components" model outperforms the ``+assignment" model, indicating that diverse assignment for boundary discrimination can effectively improve the robustness of entity span detection. Moreover, for correlation purification, adding both correlation facilitation loss and interference information filtering loss to the base model performs better than adding any single module, which demonstrates that those two modules are beneficial to improve the adversarial robustness and can boost each other.

Furthermore, the adversarial robustness is significantly improved after adding all the above modules, that is, our BDCP model performs best. 
It illustrates that when handling adversarial examples in few-shot NER, our BDCP model can prevent the cross-domain transfer learning from rote memorizing the surface features of entities and exploiting preferences in the data.

\begin{table}[h!]
	\newcommand{\tabincell}[2]{\begin{tabular}{@{}#1@{}}#2\end{tabular}}
	\centering
	\caption{Ablation study results on Few-NERD dataset.}
	\label{table_4_3}
	\begin{tabular}{lc|c|c|c}
		\toprule
		\multirow{2}{*}{Models } & \multicolumn{2}{c}{INTRA} & \multicolumn{2}{c}{INTER}\\
		\cline{2-5}
		& 5-1 & 10-5 & 5-1 & 10-5 \\
		\midrule
		Base  & 33.73 & 45.16 & 49.36 & 54.68 \\
		\midrule
		+ assignment   & 34.86 & 45.87 & 51.03 & 55.67  \\ 
		+ components  & 36.25 & 47.24 & 52.58 & 57.26\\
		\midrule
		+ facilitating  & 33.81 & 46.20 & 50.95 & 55.53  \\ 
		+ filter  & 35.38 & 47.66 & 53.42 & 58.70  \\ 
		+ purify  & 37.41 & 48.12 & 54.34 & 60.19  \\ 
		\midrule
		BDCP & \textbf{40.76} & \textbf{50.71} & \textbf{56.84} & \textbf{64.97}  \\ 
		\bottomrule 
	\end{tabular}
\end{table}

\subsection{Visualization}
\begin{figure}[!h]
	\centering
	\subfigure[Decomposed]{
		\includegraphics[width=3.5cm]{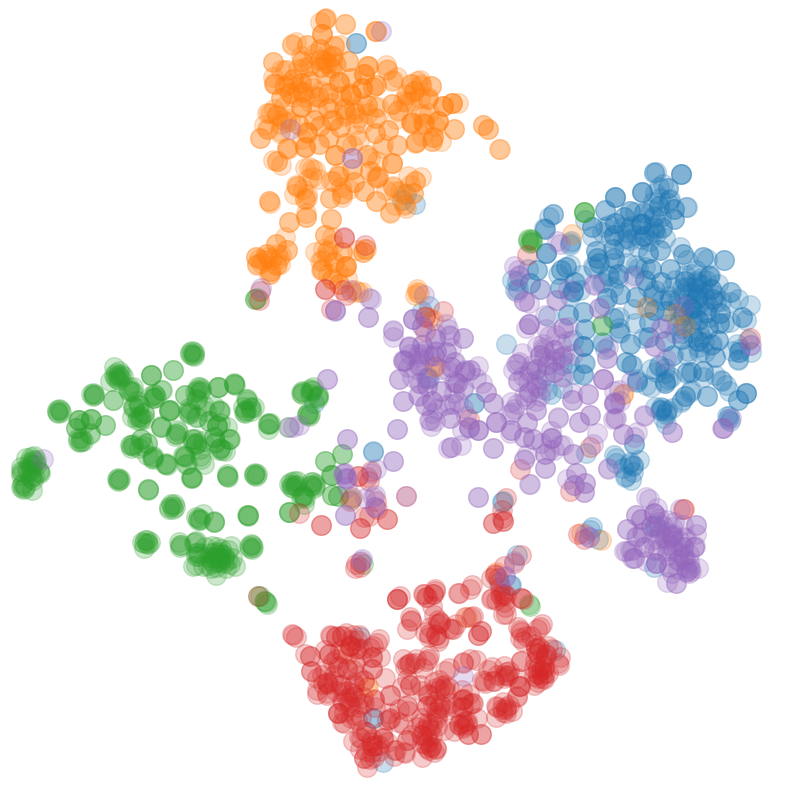}
	}
	\subfigure[Our BDCP]{
		\includegraphics[width=3.5cm]{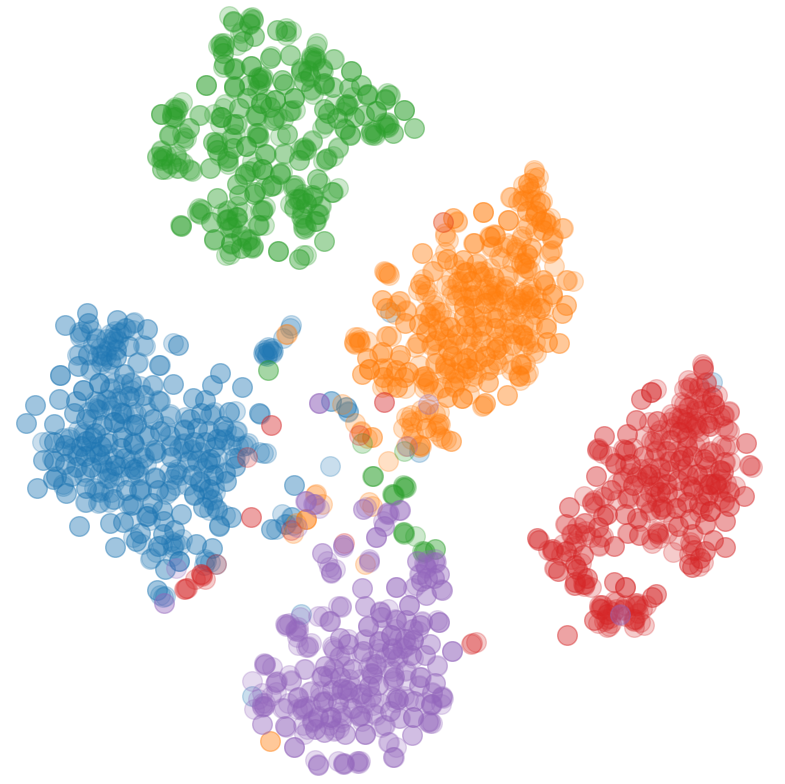}
	}
	\caption{t-SNE Visualization of span representations on Few-NERD INTER $5$-way $1$-shot query set that containing adversarial examples. The representations are obtained from Decomposed \cite{decomposed} model and our BDCP model, respectively. Different colors represent different entity types.}
	\label{fig_4_1}
\end{figure}

In order to intuitively illustrate the effectiveness of boundary discrimination and correlation purification, t-SNE \cite{van2008visualizing} is used to reduce the dimensionality of the span representations from the BDCP model and the Decomposed model, and visualize the span representations of different entity types, as shown in Figure \ref{fig_4_1}.

It can be observed that when handling textual adversarial examples in few-shot NER, the span representations of different entity types generated by the Decomposed model are less discriminative than the BDCP model. Additionally, there are more outlier span representations for entity typing. In contrast, the BDCP model can better disperse the span representations of different entity types, and has a clearer decision boundary than Decomposed model. Therefore, the BDCP model can provide textual representation space with better adversarial robustness and generalization for few-shot NER through boundary discrimination and correlation purification, under textual adversarial attack scenario.

\section{Conclusion}
In this paper, we have explored and evaluated the adversarial robustness of cross-domain transfer learning ability in few-shot NER for the first time. 
Extensive experiments indicate that existing few-shot NER methods are vulnerable when handling textual adversarial examples. 
To address this issue, we propose a robust few-Shot NER method with boundary discrimination and correlation purification. 
Specifically, in the span detection stage, the entity boundary discriminative module is introduced to diversify the assignment of token representations to corresponding closest components, thus providing a highly discriminative boundary representation space. 
In the entity typing stage, correlation purification is performed by minimizing interference information and facilitating correlation generalization to alleviate the perturbations caused by textual adversarial attacks. 
Experiments on various few-shot NER datasets containing adversarial examples demonstrate the adversarial robustness of the BDCP model. 
In future work, we will explore the robustness of adversarial training in few-shot NER.

\section{Acknowledgments}
The work is supported by the National Natural Science Foundation of China (62072039). 
We thank the anonymous reviewers for their valuable comments and feedback.

\bibliography{aaai24}

\section{Appendix}

\subsection{Preliminaries}

\subsubsection{Information Bottleneck.} 
Directly applying Information Bottleneck (IB) to few-shot NER can not improve the textual adversarial robustness \cite{WangDXZZG0CH22}. This is because not only noise and outliers exist in the textual adversarial scenarios, but also few-shot NER model focuses on the cross-domain transfer ability of that model. On the one hand, current research has demonstrated that it is challenging to trade-off high compression and accurate prediction \cite{TishbyS00,WangBLTZ19,Piran2020}. Consequently, when compressing noise, the generalization information that is crucial to the cross-domain transfer learning ability is inevitably left out. 
In addition, directly applying IB results in the neural networks exploiting accessible shortcut information to solely maximize prediction accuracy \cite{IlyasSTETM19,WangDXZZG0CH22}. It makes the few-shot NER model memorize entity names instead of recognizing unseen entities according to the correlations between entities and contexts \cite{AgarwalYWN21}. We introduce how to extend IB to improve the textual adversarial robustness in few-shot NER.

According to the chain rule of mutual information \cite{Federici0FKA20,WangDXZZG0CH22}, $I(X;T)$ can be decomposed into two terms:
\begin{equation}
	I(X;T)=\underbrace{I(T;Y)}_{predictive}+\underbrace{I(X;T|Y)}_{redundant},
	\label{equ_3_11}
\end{equation}

\noindent where the first term $I(T;Y)$ denotes the information in $T$ that is accessible to predict $Y$. 
The second term $I(X;T|Y)$ represents the redundant information in $T$, which is not predictive of $Y$.

\subsection{Methodology}

\subsubsection{Class Prototypes.} 
The prototypical network ProtoNet \cite{SnellSZ17} is leveraged as the backbone model in correlation purified entity typing stage. 
For each entity class $y_k$ in the support set, the prototype $c_k$ for each entity category $y_k$ is computed by averaging all span representations $\tilde{h}_{s,k}$ belonging to the entity class $y_k$:

\begin{equation}
	{{c}_{k}}=\frac{1}{\left| {{S}_{k}} \right|}\sum\limits_{{{{\tilde{h}}}_{s,k}}\in {{S}_{k}}}{{{{\tilde{h}}}_{s,k}}},
	\label{equ_3_ck}
\end{equation}

\noindent where $S_k$ represents the set of entity spans belonging to the entity class $y_k$. 
$\left| {{S}_{k}} \right|$ indicates the number of entity spans in $S_k$.

\subsubsection{Prototype Loss.} 
During training, the prototypes of all entity classes $Y_train$ in the support set ${\mathcal{S}_{train}}$ are calculated by Eq. (\ref{equ_3_ck}). 
Next, according to the distance between the entity span ${\tilde{h}}_s$ and the prototype $c_k$ of the entity class $y_k$ in the query set ${\mathcal{Q}_{train}}$, the distribution probability of the entity classes for entity span ${\tilde{h}}_s$ is calculated as:

\begin{equation}
	{{p}_{c}}({{y}_{k}};{{\tilde{h}}_{s}})=\frac{\exp (-d({{c}_{k}},{{{\tilde{h}}}_{s}}))}{\sum\limits_{{{y}_{i}}\in {{Y}_{train}}}{\exp (-d({{c}_{i}},{{{\tilde{h}}}_{s}}))}},
	\label{equ_3_pyk}
\end{equation}

\noindent where $d(\cdot ,\cdot )$ denotes the distance function. 
The prototype loss is optimized by using the cross-entropy loss:

\begin{equation}
	{\mathcal{L}_{t}}=\sum\limits_{{{{\tilde{h}}}_{s}}\in {{Q}_{train}}}{-\log {{p}_{c}}({{y}_{s}};{{{\tilde{h}}}_{s}})},
	\label{equ_3_lt}
\end{equation}

\noindent where $y_s$ represents the ground-truth entity class of entity span $\tilde{h}_s$.

\subsubsection{Correlation Facilitating.} 

For the correlation facilitating in entity typing stage, $I(\tilde{h}_s,\tilde{h}_c)$ is a lower bound of $I(\tilde{h}_s,{x}')$.
\begin{equation}
	\begin{aligned}
		I({{\tilde{h}}_{s}};{x}')& =I({{\tilde{h}}_{s}};{x}'{{\tilde{h}}_{c}})-I({{\tilde{h}}_{s}};{{\tilde{h}}_{c}}|{x}') \\
		& =I({{\tilde{h}}_{s}};{x}'{{\tilde{h}}_{c}}) \\
		& =I({{\tilde{h}}_{s}};{{\tilde{h}}_{c}})+I({{\tilde{h}}_{s}};{x}'|{{\tilde{h}}_{c}}) \\
		& \ge I({{\tilde{h}}_{s}};{{\tilde{h}}_{c}})
		\label{equ_app_max}
	\end{aligned}
\end{equation}

\subsubsection{Correlation Purification.} 
For the correlation purification in entity typing stage, the upper bound of $I({x};{\tilde{h}_s}|{{x}'})$ is:
\begin{equation}
	\begin{aligned}
		& I(x;{{\tilde{h}}_{s}}|{x}') \\
		& ={{E}_{x,{x}'\sim p(x,{x}')}}{{E}_{t\sim p({{{\tilde{h}}}_{s}}|{{v}_{s}})}}\log \frac{p(x,{{{\tilde{h}}}_{s}}|{x}')}{p(x|{x}')p({{{\tilde{h}}}_{s}}|{x}')} \\
		& ={{E}_{x,{x}'\sim p(x,{x}')}}{{E}_{t\sim p({{{\tilde{h}}}_{s}}|{{v}_{s}})}}\log \frac{p({{{\tilde{h}}}_{s}}|x)p(x|{x}')}{p(x|{x}')p({{{\tilde{h}}}_{s}}|{x}')} \\
		& ={{E}_{x,{x}'\sim p(x,{x}')}}{{E}_{t\sim p({{{\tilde{h}}}_{s}}|{{v}_{s}})}}\log \frac{p({{{\tilde{h}}}_{s}}|x)}{p({{{\tilde{h}}}_{s}}|{x}')} \\
		& ={{E}_{x,{x}'\sim p(x,{x}')}}{{E}_{t\sim p({{{\tilde{h}}}_{s}}|{{v}_{s}})}}\log \frac{p({{{\tilde{h}}}_{s}}|x)p({{{\tilde{h}}}_{c}}|{x}')}{p({{{\tilde{h}}}_{c}}|{x}')p({{{\tilde{h}}}_{s}}|{x}')} \\
		& ={{D}_{KL}}(p({{\tilde{h}}_{s}}|x)||p({{\tilde{h}}_{c}}|{x}'))-{{D}_{KL}}(p({{\tilde{h}}_{s}}|{x}')||p({{\tilde{h}}_{c}}|{x}')) \\
		& \le {{D}_{KL}}(p({{\tilde{h}}_{s}}|x)||p({{\tilde{h}}_{c}}|{x}')).
		\label{equ_app_min}
	\end{aligned}
\end{equation}

However, under the textual adversarial attack scenario, only the attacked example ${x}'$ is available, while the original example $x$ is not available. 
To explicitly minimize the redundant information in the correlations between entities and contexts, we attempt to find the internal representations $t_s$ and $t_c$, and replace the upper bound in Eq. \ref{equ_app_min} with ${{D}_{KL}}({{p}}({{t}_{s}}|{{\tilde{h}}_{s}})||{{p}}({{t}_{c}}|{{\tilde{h}}_{c}})))$. 
The performance of the ``+filter" model in the Ablation Study section shows the effectiveness of this operation.

\subsubsection{Adversarial Examples Generating.} 
Bert-Attack \cite{LiMGXQ20} was originally used for text classification and sentiment analysis tasks. We use Bert-Attack to generate adversarial examples for the few-shot NER task by modifying the data format in Few-NERD and Cross-Dataset datasets. We treat entity types in sentences as labels, then use Bert-Attack to find the vulnerable words in sentences and replace them with synonyms.


\subsection{Experiments}

\subsubsection{Baselines.}

We compare the proposed BDCP model with the following typical metric-learning based baselines:

\begin{enumerate}[1)]
	\item \textbf{ProtoBERT} \cite{FritzlerLK19} utilizes a token-level prototypical network which represents each class by averaging token	representation with the same label, then the label of each token in the query set is decided by its nearest class prototype.
	\item \textbf{StructShot} \cite{YangK20} pretrains BERT for token embedding by conventional classification for training, and uses an abstract transition probability for Viterbi decoding at testing.
	\item \textbf{CONTAINER} \cite{DasKPZ22} uses token-level contrastive learning for training BERT as token embedding function, then finetune the BERT on support set and apply a nearest neighbor method at inference time.
	\item \textbf{Decomposed} \cite{decomposed} addresses the problem of few-shot NER by sequentially tackling few-shot span detection and few-shot entity typing using meta-learning.
	\item \textbf{MatchingNet} \cite{VinyalsBLKW16} is similar to ProtoBERT except that it calculates the similarity between query instances and support instances instead of class prototypes.
	\item \textbf{SimBERT} \cite{HouCLZLLL20} applies BERT without any finetuning as the embedding function, then assign each token’s label by retrieving the most similar token in the support set.
	\item \textbf{L-TapNet+CDT} \cite{HouCLZLLL20} enhances	TapNet \cite{YoonSM19} with pair-wise embedding, label semantic, and CDT transition mechanism.
\end{enumerate}

In addition, we compare with direct adaptations of the following textual defense methods: RockNER \cite{LinG0M021}, RanMASK \cite{ZengXZH23} for both Few-NERD and Cross-Dataset\footnote{We apply data augmentation and random masking to Decomposed, using the code from https://github.com/microsoft/vert-papers/tree/master/papers/DecomposedMetaNER.}:

\begin{enumerate}[1)]
	\item \textbf{RockNER} \cite{LinG0M021} replace each entity in the target sentence with a different entity of the same type from another sentence to improve the robustness of named entity recognition.
	\item \textbf{RanMASK} \cite{ZengXZH23} repeatedly performs random masking operations on the input sentences in order to generate a large set of masked copies of the sentences.
\end{enumerate}

\noindent \textbf{Implementation Details.} 
$\mathbb{E}_{1}$ and $\mathbb{E}_{2}$ are $Softplus$ and $ReLU$ activation functions, respectively.

\noindent \textbf{About Large Language Model.} We have used the large language model, namely, ChatGPT-GPT-3.5 for few-shot NER. The training corpus of GPT-3.5 are likely to contain those two datasets in this work. For fair comparison, the sentences in a medical entity recognition dataset we constructed are input into the ChatGPT to perform few-shot NER. We obtain the following results: The F1 scores of GPT-3.5 is 2.87 (63.34-60.47) and 1.23 (54.85-53.62) lower than our BDCP on clean and attacked examples, and the reason is that the recall of GPT-3.5 is lower than that of our BDCP.

\noindent \textbf{About Traditional NER.} We evaluate the BDCP model on traditional NER task. Actually, in experiments, the traditional NER and Few-shot NER can be converted via the domains of the samples. To perform traditional NER task, we have adjusted the episodes construction of Few-NERD INTER dataset so that the domains of the validation set and test set are the same as those of the training set. We construct a new baseline called BDCP-T by removing boundary discrimination and correlation purification. The F1 scores of the BDCP and BDCP-T models on traditional NER task are 87.63 and 80.27 points, respectively. It can be observed that our BDCP can achieve considerable performance on the traditional NER task.

\noindent \textbf{About Performance Analysis.} To analyze the performance improvement of different types of adversarial examples, we add adversarial examples of entity-level attacks, which are generated by replacing entities with other entities of the same fine-grained class in Wikidata. The results shows that the BDCP model has better performance improvement against the entity-level attack (56.84 and 59.37 F1 scores for random synonym substitution and entity-level attack), because the BDCP model can learn robust entity information from the semantic association between entities and contexts.

\noindent \textbf{About One Sequence Labeling.} Accordingly, we have modified the BDCP model to a single-stage approach by adopting one sequence labeling module, called BDCP-Single. That is, we remove the entity typing stage, and modify the classification layer to the sequence labeling layer. In 1-shot setup of the Few-NERD INTRA, the F1 scores of BDCP-Single model on clean examples and attacked examples are 39.58 and 24.14 points respectively. Compared with the BDCP model, the F1 scores decrease by 13.05 and 16.62 points respectively. This is because the two-stage model can adequately explore the information brought by the support examples from more perspectives such as span boundary and correlation between entities and contexts.

\subsection{Limitations and Future Work}

\textbf{Limitations}: Two-stage paradigm may lead to error propagation. The reason lies in that span detection in the first stage is the prerequisite for entity typing in the second stage.

\noindent \textbf{Future Work}: Within our two-stage paradigm, one future work is that the adversarial training can be used to improve the cross-domain learning capability of few-shot NER. Another work is that both clean and attacked examples can be used for contrastive learning to achieve better performance.

\end{document}